\pdfoutput=1
\documentclass[10pt,twocolumn,letterpaper]{article}
\usepackage[pagenumbers]{cvpr}
\usepackage{xspace}
\usepackage{microtype}
\usepackage{multirow}
\usepackage{array}
\usepackage{tabularx}

\definecolor{cvprblue}{rgb}{0.15,0.38,0.62}
\usepackage[pagebackref,breaklinks,colorlinks,allcolors=cvprblue]{hyperref}

\newcommand{\method}{AlignHand\xspace}
\newcommand{\R}{\mathbb{R}}

\hypersetup{
    pdfauthor={Zhilu Yang, Kun Yang, Shiyan Wang, Anqi Xu, Xinran Huang, Yizheng Jiao, Huan Yang, Rolandos Alexandros Potamias},
    pdftitle={AlignHand: Multi-View Hand Reconstruction from a Frozen Monocular Prior}
}

\title{AlignHand: Multi-View Hand Reconstruction\\from a Frozen Monocular Prior}

\author{
Zhilu Yang$^{1}$\quad
Kun Yang$^{2}$\quad
Shiyan Wang$^{3}$\quad
Anqi Xu$^{4}$\quad
Xinran Huang$^{5}$\\
Yizheng Jiao$^{6}$\quad
Huan Yang$^{7}$\quad
Rolandos Alexandros Potamias$^{1*}$\\[0.1em]
{\small $^{1}$Imperial College London\quad
$^{2}$Beihang University\quad
$^{3}$New York University Abu Dhabi\quad
$^{4}$Hong Kong Baptist University}\\
{\small $^{5}$The Chinese University of Hong Kong, Shenzhen\quad
$^{6}$ByteDance\quad
$^{7}$South China Normal University}\\
\href{https://ada-zy2425.github.io/AlignHand/}{\texttt{ada-zy2425.github.io/AlignHand}}
}

\begin{document}

\twocolumn[{\renewcommand\twocolumn[1][]{#1}\maketitle

\begin{center}
\includegraphics[width=0.95\textwidth]{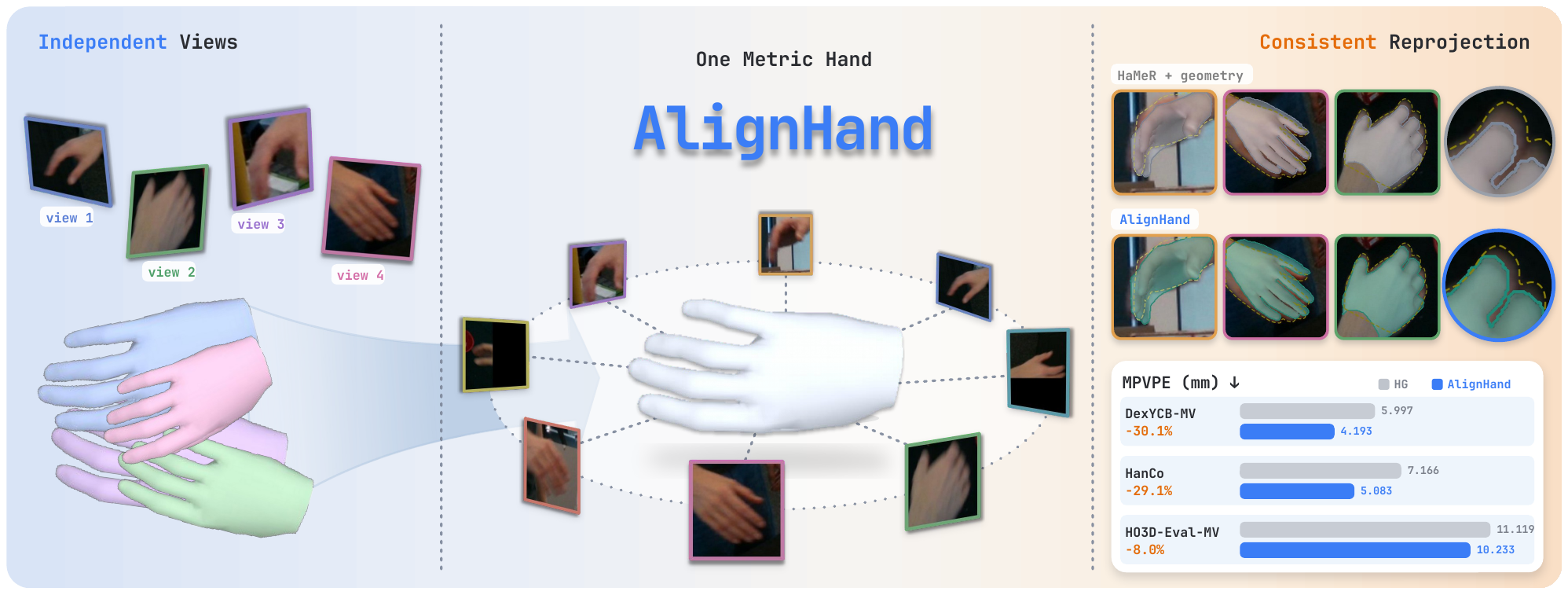}
\captionof{figure}{\textbf{Overview of AlignHand.} Given calibrated multi-view images, AlignHand aligns the independent predictions of a frozen monocular estimator in a shared metric frame and integrates their complementary surface evidence into a consistent hand reconstruction, reducing MPVPE relative to the HaMeR + geometry baseline (HG) on DexYCB-MV, HanCo, and HO3D-Eval-MV.}
\label{fig:teaser}
\end{center}
\vspace{0.6em}}]

\begingroup
\renewcommand{\thefootnote}{\fnsymbol{footnote}}
\footnotetext[1]{Corresponding author.}
\endgroup
\begin{abstract}
Modern monocular hand estimators recover detailed 3D meshes from individual images, yet effectively leveraging their predictions across multiple views remains challenging. Independently predicted meshes can disagree in scale, placement, and articulation, limiting the integration of complementary multi-view information. Hence, we present AlignHand, a calibrated multi-view framework that builds on a frozen monocular hand reconstruction prior to align and enhance its predictions. Specifically, AlignHand refines cross-view joint observations and estimates their reliability to establish a shared metric reference for alignment. Once the meshes are brought into a common frame, their complementary surface evidence is selectively fused across anatomical regions, followed by pixel-aligned refinement that predicts bounded residual displacements from local image features. This global-to-local integration enables AlignHand to exploit complementary observations while preserving the strong monocular prior. Extensive experiments on DexYCB-MV, HanCo, and HO3D-Eval-MV demonstrate that AlignHand achieves the lowest MPVPE and MPJPE among the compared methods under the respective evaluation protocols, reducing MPVPE by 30.1\%, 29.1\%, and 8.0\%, respectively, compared with a matched geometry baseline.
\end{abstract}

\section{Introduction}
\label{sec:introduction}

Multi-view hand reconstruction aims to recover hand shape, articulation, and metric placement from synchronized images. Recent monocular hand estimators recover detailed meshes from individual images~\cite{pavlakos2024hamer,potamias2025wilor}, providing strong geometric priors for 3D hand reconstruction. Multiple calibrated views provide complementary visual evidence, as different viewing angles can reveal hand regions occluded in other views. However, applying a monocular estimator independently to each view does not guarantee a consistent multi-view reconstruction. The resulting meshes may disagree in metric scale, placement, and local articulation, even when each prediction appears plausible in its source image. Surface quality also varies across views and hand regions. Although camera calibration specifies the relationships between views, it does not by itself resolve inconsistencies among independently predicted meshes. Effectively exploiting these monocular predictions therefore requires consistent cross-view alignment and integration of their complementary surface information.

Existing multi-view hand reconstruction methods exploit cross-view information through different strategies. For multi-view inference, POEM and POEMv2 aggregate image features around 3D points~\cite{yang2023poem,yang2025poemv2}, while MLPHand combines triangulated joints and multi-view features with a frozen skeleton-to-mesh model~\cite{yang2024mlphand}. JUMP-Hand further employs joint-wise uncertainty to guide view fusion~\cite{kuang2026jumphand}. Multi-view information can also improve monocular learning, as demonstrated by SiMA-Hand~\cite{wang2024simahand}. However, integrating independently decoded monocular meshes presents three challenges: (i) Noisy 2D joint observations weaken cross-view geometric consistency. (ii) Independently predicted meshes exhibit inconsistent metric scale and placement. (iii) Region-dependent reconstruction quality cannot be adequately captured by a single mesh-level fusion weight.

To address these challenges, we present \method, a calibrated multi-view framework that builds on a frozen monocular reconstruction prior to align and enhance its predictions. Specifically, (i) cross-view observation refinement corrects predicted 2D joints and estimates their view-dependent reliability, providing more consistent geometric evidence across cameras. (ii) Support-weighted triangulation constructs a shared metric skeleton, which guides robust alignment of independently predicted meshes into a common coordinate frame. (iii) Region-specific fusion integrates complementary surface evidence from the aligned meshes, followed by bounded pixel-aligned residual prediction using local image features. The view support estimated during observation refinement informs both metric alignment and subsequent surface integration. Together, these components enable \method to effectively exploit complementary multi-view information and improve hand reconstruction from strong monocular priors.

\section{Related Work}
\label{sec:related}

\begin{figure*}[t]
\centering
\includegraphics[width=\textwidth]{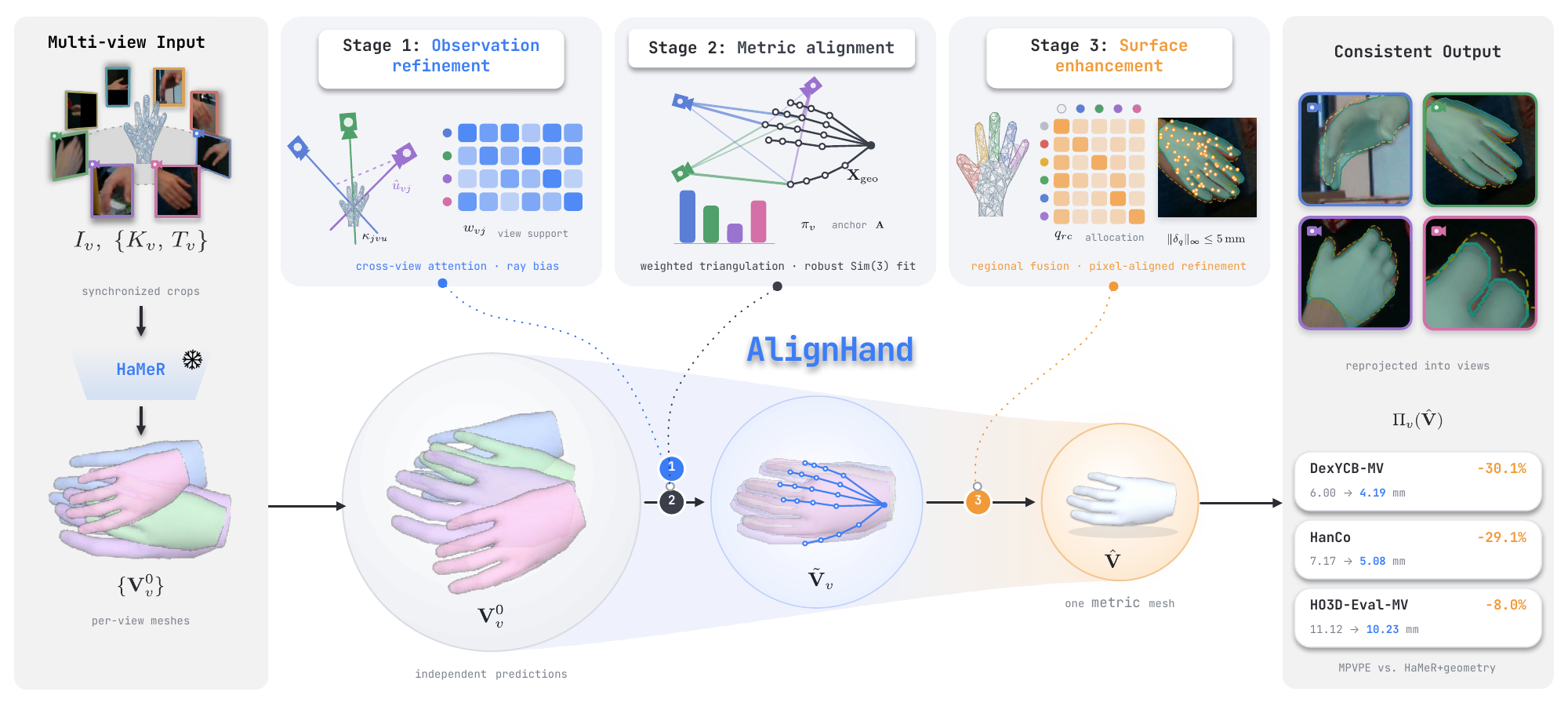}
\vspace{-0.5cm}
\caption{\textbf{Aligning and enhancing the outputs of one frozen monocular prior.} Observation refinement corrects 2D joints and estimates view support, metric alignment fits each decoded mesh to the weighted triangulated skeleton, and surface enhancement integrates the aligned candidates into the final mesh.}
\label{fig:architecture}
\end{figure*}

\vspace{0.2cm}
\noindent\textbf{Monocular Hand Mesh Reconstruction.}
Monocular hand reconstruction has progressed from lifting 2D keypoints to recovering detailed 3D meshes~\cite{zimmermann2017learning,ge2019handshape}. Existing approaches can be broadly grouped into two categories. (i) Parametric methods predict hand shape and articulation using structured models such as MANO~\cite{romero2017mano,boukhayma2019wild}. HaMeR leverages large-scale transformer-based learning for detailed hand mesh recovery, while WiLoR integrates hand localization with high-fidelity mesh estimation~\cite{pavlakos2024hamer,potamias2025wilor}. Their structured mesh outputs provide consistent vertex correspondence, facilitating geometric processing of independently estimated surfaces. (ii) Direct mesh methods infer surface geometry through learned vertex representations. METRO, MeshGraphormer, and CMR exploit attention, mesh connectivity, and semantic aggregation~\cite{lin2021metro,lin2021meshgraphormer,chen2021cmr}. I2L-MeshNet predicts mesh vertices through lixel heatmaps, while Pose2Mesh recovers meshes from joint representations using graph convolutions~\cite{moon2020i2lmeshnet,choi2020pose2mesh}. HandOccNet and MobRecon address occlusion robustness and efficient inference, respectively~\cite{park2022handoccnet,chen2022mobrecon}. HandOS unifies hand detection, 2D pose estimation, and 3D mesh reconstruction in a one-stage framework built on a frozen detector~\cite{chen2025handos}. Complementary to these architectural advances, AnyHand provides large-scale synthetic RGB(-D) data that improves existing RGB-based estimators, including HaMeR and WiLoR, without changing their architectures~\cite{si2026anyhand}.

Related research further enhances learned 3D predictions through geometric fitting and image-aligned refinement. SMPLify fits a parametric body model to image keypoints, while SPIN incorporates model fitting into regression-based learning~\cite{bogo2016smplify,kolotouros2019spin}. Another direction connects image evidence with predicted 3D surfaces. PIFu uses pixel-aligned features for implicit surface reconstruction, whereas PyMAF samples mesh-aligned features to iteratively refine predicted model parameters~\cite{saito2019pifu,zhang2021pymaf}. However, refining individual predictions does not directly establish consistency among meshes independently estimated from different camera views.

\vspace{0.2cm}
\noindent\textbf{Multi-View Hand Reconstruction.}
Multi-view reconstruction exploits complementary observations through geometric constraints and learned cross-view interaction. (i) Geometry-based methods recover 3D structure from calibrated image observations. Classical triangulation uses camera projection constraints, while learnable triangulation incorporates view-dependent confidence to account for unequal observation quality~\cite{hartley1997triangulation,iskakov2019learnable}. (ii) Feature-based methods combine visual evidence through cross-view feature exchange, epipolar aggregation, and adaptive view weighting, as demonstrated by Cross View Fusion, Epipolar Transformers, and AdaFuse~\cite{qiu2019crossview,he2020epipolar,zhang2021adafuse}. These techniques also support hand-specific reconstruction. For instance, POEM and POEMv2 aggregate image features around 3D points to recover hand meshes~\cite{yang2023poem,yang2025poemv2}. JUMP-Hand uses joint-wise uncertainty to adaptively gate view experts~\cite{kuang2026jumphand}. MLPHand feeds triangulated joints and multi-view features to a pretrained skeleton-to-mesh model~\cite{yang2024mlphand}, whereas we integrate meshes already decoded by a frozen image-to-mesh estimator.

Beyond direct multi-view inference, other studies use additional observations to improve hand estimators or extend reconstruction to different settings. SiMA-Hand uses multi-view features to guide monocular learning~\cite{wang2024simahand}, HaMuCo exploits collaborative cross-view self-supervision~\cite{zheng2023hamuco}, and S2DHand adapts a pretrained single-view estimator to dual-view inputs~\cite{liu2024singledual}. Temporal approaches such as UmeTrack and UST-Hand incorporate observations across frames for hand tracking and reconstruction~\cite{han2022umetrack,han2026usthand}, while HGGT jointly estimates hand meshes and cameras from uncalibrated images~\cite{liu2026hggt}. However, independently decoded monocular meshes may retain inconsistent metric placement and region-dependent errors, making their complementary information difficult to integrate. In contrast, \method aligns these predictions in a shared metric frame and integrates their complementary surface evidence while keeping the monocular estimator frozen.

\section{Methodology}
\label{sec:method}

\subsection{Problem Formulation}
\label{sec:formulation}

Given $V$ synchronized hand images
$\mathcal I=\{I_v\}_{v=0}^{V-1}$ and their camera
intrinsics and extrinsics $\{K_v,T_v\}$, our goal is
to reconstruct a single metric hand mesh
$\widehat{\mathbf V}\in\R^{778\times3}$
in the coordinate frame of a reference camera.
The mesh follows MANO topology~\cite{romero2017mano}, and its joints are obtained
through a fixed mesh-to-joint operator
$\widehat{\mathbf X}=\mathcal J(\widehat{\mathbf V})
\in\R^{21\times3}$.
The reconstruction builds on independently decoded
monocular predictions and exploits complementary
observations across calibrated views.
We assume prelocalized hand inputs.
No ground-truth 3D joints or meshes
are used during inference.

\subsection{Overall Framework}
\label{sec:framework}

Fig.~\ref{fig:architecture} illustrates \method,
which integrates predictions from a frozen monocular
image-to-mesh estimator through three successive
components. For each view $v$, HaMeR produces a
local mesh $\mathbf V^0_v$, 2D joints
$\mathbf u^0_{vj}$, and image features $\Phi_v$.
(i) Cross-view observation refinement corrects
the predicted 2D joints and estimates joint-specific
view support $w_{vj}$ to obtain reliable geometric
evidence across cameras.
(ii) Metric alignment uses the refined observations
and their support to triangulate a shared metric
skeleton $\mathbf X_{\mathrm{geo}}$, aligns each
decoded mesh to obtain $\widetilde{\mathbf V}_v$,
and combines the candidates into a fit-aware
anchor $\mathbf A$.
(iii) Surface enhancement integrates complementary
information from the aligned candidates through
region-specific fusion to obtain
$\mathbf V^{\mathrm{fus}}$, followed by bounded
pixel-aligned refinement to produce the final mesh
$\widehat{\mathbf V}$.
The estimated view support is reused throughout
alignment and surface enhancement, connecting
cross-view geometric reliability with subsequent
local reconstruction.

\subsection{Cross-View Observation Refinement}
\label{sec:evidence_refinement}

Calibrated triangulation recovers a common 3D
reference from multi-view joint correspondences~\cite{hartley1997triangulation,iskakov2019learnable}.
Here, however, these correspondences are predicted
independently by the frozen monocular estimator.
A mislocalized joint changes its viewing ray,
while poor visibility can make observations
of the same joint unequally reliable.
Such errors propagate into the triangulated
skeleton and subsequently affect the alignment
of the decoded meshes.
Coordinate correction and view support address
these problems differently: the former adjusts
the observed joint locations, whereas the latter
controls their influence on shared geometry.
We therefore jointly refine 2D coordinates and
estimate joint-specific view support from
calibrated rays and image features.

Specifically, for joint $j$ in view $v$, we sample
an appearance descriptor $\mathbf a_{vj}$ from
$\Phi_v$ at the initial joint $\mathbf u^0_{vj}$,
given in normalized crop coordinates within
$[-0.5,0.5]^2$. The crop mapping $\mathcal C_v$
converts these coordinates to image pixels.
Given $K_v$ and the reference-to-camera transform
$T_v=[R_v^{\mathrm{cam}}\mid
\mathbf t_v^{\mathrm{cam}}]$, each initial
observation defines a ray with unit direction
$\mathbf d_{vj}$ and moment
$\mathbf m_{vj}=\mathbf c_v\times\mathbf d_{vj}$
in Pl\"ucker coordinates~\cite{sitzmann2021lfn,zhang2024raydiffusion},
where $\mathbf c_v=-(R_v^{\mathrm{cam}})^\top\mathbf t_v^{\mathrm{cam}}$
is the camera center.
We construct a joint token combining appearance,
ray geometry, and relative camera pose:
\[
\mathbf t_{vj}
=W_a\mathbf a_{vj}
+W_g[\mathbf u^0_{vj},\mathbf d_{vj},
\mathbf m_{vj},\mathbf p_v]
+\mathbf e_j,
\]
where $[\cdot]$ denotes concatenation (17-D),
$W_a,W_g$ are learned projections,
$\mathbf e_j$ encodes joint identity, and
$\mathbf p_v\in\R^9$ represents the camera pose
relative to the reference frame. It comprises
the first two columns of
$(R_v^{\mathrm{cam}})^\top$ and the camera center normalized
by the mean distance of available cameras from
the reference camera.
Cross-view attention~\cite{vaswani2017attention}
compares tokens corresponding to the same joint.
To encourage geometric compatibility, we compute
the ray discrepancy
\[
\kappa_{jvu}
=\left|
\mathbf d_{vj}^{\top}\mathbf m_{uj}
+\mathbf d_{uj}^{\top}\mathbf m_{vj}
\right|,
\]
which vanishes for rays passing through a common
3D point. The attention bias
$-\min(\kappa_{jvu}/s_j,12)$ uses the median
discrepancy $s_j$, computed without gradients
and floored at $10^{-5}$, to favor geometrically
compatible observations.
The resulting token $\mathbf h_{vj}$
is passed to two prediction heads with
zero-initialized output layers:
\[
\begin{aligned}
\widehat{\mathbf u}_{vj}
&=\mathbf u^0_{vj}
+0.05\tanh f_{\mathrm{corr}}(\mathbf h_{vj}),\\
w_{vj}
&=\underset{v\in\mathcal V}{\operatorname{softmax}}
\left(-\frac{e_{vj}}{5}
+0.25f_{\mathrm{sup}}(\mathbf h_{vj})\right),
\end{aligned}
\]
where the coordinate correction is bounded by
12.8 crop pixels per coordinate.
The geometric term
$e_{vj}=\|\Pi_v(\mathbf X_j^{\mathrm{uni}})
-\widehat{\mathbf p}_{vj}\|_2$
measures the pixel reprojection error of the
corrected observation
$\widehat{\mathbf p}_{vj}
=\mathcal C_v(\widehat{\mathbf u}_{vj})$
relative to a provisional uniform triangulation
$\mathbf X_j^{\mathrm{uni}}$ computed from
the corrected observations, and $\Pi_v$
denotes projection into view $v$.
The support combines measured geometric
disagreement with learned appearance cues,
with unavailable views masked.
Zero initialization preserves the original joint
coordinates and initially determines support
from geometric reprojection errors alone.
During training, the corrected observations and
predicted support jointly determine the metric
skeleton $\mathbf X_{\mathrm{geo}}$, allowing
both heads to be supervised through the geometry
they produce. The observation objective is
\[
\begin{aligned}
\mathcal L_{\mathrm{obs}}
={}&
\left(S_{0.01}^{3D}+0.25E^{3D}\right)
+\left(0.5S_{0.02}^{2D}+0.25S_{0.02}^{\mathrm{proj}}\right)\\
&+0.01\left(\mathcal L_H+\mathcal L_{\Delta u}\right).
\end{aligned}
\]
Here $S_\beta$ denotes the mean smooth-$L_1$
loss~\cite{girshick2015fastrcnn} with transition
$\beta$. The terms $S_{0.01}^{3D}$ and
the mean joint error $E^{3D}$ supervise the
metric skeleton in meters.
$S_{0.02}^{2D}$ supervises the corrected
image joints against annotations, whereas
$S_{0.02}^{\mathrm{proj}}$ enforces consistency
between skeleton reprojections and corrected
joints. Both operate in normalized crop
coordinates over supported annotations.
The entropy penalty $\mathcal L_H$ discourages
prematurely concentrated support by penalizing
entropy below $H_{\min}=\min(1.2,\log|\mathcal V|)$,
and $\mathcal L_{\Delta u}$ penalizes
unnecessary coordinate corrections.

\subsection{Metric Alignment}
\label{sec:mesh_recovery}

Refined cross-view observations provide the
evidence for recovering a common metric
skeleton, but this sparse representation
does not capture the detailed surfaces
already predicted by the monocular estimator.
Conversely, the independently decoded meshes
retain local articulation while differing
in scale, orientation, and placement,
preventing their direct integration.
We therefore assign complementary roles
to multi-view geometry and the frozen prior:
triangulation establishes a shared metric
reference, while similarity fitting brings
each predicted mesh into that frame without
reconstructing its surface from scratch.
Fitting all candidates to the same reference
reduces global coordinate discrepancies while
retaining view-dependent articulation and
surface differences. Since global alignment
cannot eliminate these local differences, we further construct
a fit-aware anchor that favors meshes
consistent with the observed skeleton
before subsequent surface integration.

Specifically, for each observation,
$D_{vj}\in\R^{2\times4}$ contains the linear
projection constraints defined by the corrected
image joint $\widehat{\mathbf p}_{vj}$ and
the camera projection matrix
$P_v\in\R^{3\times4}$.
Support-weighted triangulation~\cite{hartley1997triangulation,iskakov2019learnable}
recovers
\[
\begin{aligned}
\mathbf z^\star_j
&=\underset{\|\mathbf z\|_2=1}{\arg\min}
\sum_{v\in\mathcal V}w_{vj}\|D_{vj}\mathbf z\|_2^2,\\
\mathbf X_{\mathrm{geo},j}
&=(\mathbf z^\star_j)_{1:3}/(\mathbf z^\star_j)_4.
\end{aligned}
\]
Known metric camera translations establish
the scale of the triangulated skeleton.
We retain joints with at least two valid
observations and finite dehomogenized coordinates. Each decoded mesh is then aligned to
the shared skeleton by fitting HaMeR's native
joints $\mathbf X^0_{vj}$ using a similarity
transform:
\[
\begin{aligned}
(s_v,Q_v,\mathbf t_v)
={}&
\underset{s>0,\;Q\in SO(3),\;\mathbf t\in\R^3}{\arg\min}\\
&\sum_j\omega_{vj}
\left\|sQ\mathbf X^0_{vj}+\mathbf t-\mathbf X_{\mathrm{geo},j}\right\|_2^2,
\end{aligned}
\]
following point-set similarity
fitting~\cite{umeyama1991least}.
To reduce the influence of residual articulation
differences, we use three fitting passes:
an initial support-weighted fit followed by
two Huber-style reweighting passes~\cite{huber1964robust}.
The joint weights are updated according to
$\omega_{vj}\leftarrow
w_{vj}\min(1,\tau/r_{vj})$,
where $\tau=0.012$\,m and $r_{vj}$
is the current 3D residual.
This reweighting reduces the influence
of joints whose local articulation
cannot be explained by a global
similarity transform.
The estimated transform is applied to every
vertex $m\in\{1,\dots,778\}$,
$\widetilde{\mathbf V}_{vm}
=s_vQ_v\mathbf V^0_{vm}+\mathbf t_v$,
preserving the candidate's relative
surface geometry up to uniform scale.
Although the candidates share a metric
frame, their articulation accuracy may
still differ across views.
We construct a fit-aware anchor:
\[
\begin{aligned}
a_v&=\frac{\overline w_v}{\overline e_v+0.001},\qquad
\pi_v=\frac{a_v}{\sum_{u\in\mathcal V_{\mathrm{fit}}}a_u},\\
\mathbf A_m&=\sum_{v\in\mathcal V_{\mathrm{fit}}}
\pi_v\widetilde{\mathbf V}_{vm},
\end{aligned}
\]
where $\overline w_v$ denotes mean joint
support, $\overline e_v$ is mean fitting
error in meters, and
$\mathcal V_{\mathrm{fit}}$ contains valid
positive-scale fits supported by at least
three joints.
Support measures observation reliability,
whereas fitting error measures agreement
with the shared metric skeleton.
A mesh from a well-supported view may still contain
articulation errors, and a small residual
alone need not imply reliable observations.
Their combination therefore favors
candidates that are both supported and
geometrically compatible.
The anchor $\mathbf A$ provides a common
mesh-level reference, while the aligned
meshes $\widetilde{\mathbf V}_v$,
skeleton $\mathbf X_{\mathrm{geo}}$, and
support $w_{vj}$ remain available for
regional surface enhancement.

\subsection{Surface Enhancement}
\label{sec:regional_fusion}

\vspace{0.2cm}
\noindent\textbf{Regional fusion.}
Similarity alignment reduces global discrepancies
without correcting articulation errors within
individual monocular predictions.
Consequently, aligned candidates need not be
equally accurate across anatomical regions:
one view may better recover the palm, while
another more accurately captures a finger.
A single weight applied to an entire mesh
cannot express these regional preferences.
We therefore learn separate candidate
allocations for the palm and each finger,
using the fit-aware anchor as a common
geometric reference. This enables
region-dependent integration once the
candidates share a metric frame.

Specifically, the candidate bank is
$\mathcal C=\{\mathbf A\}\cup
\{\widetilde{\mathbf V}_v\}_{v\in\mathcal V_{\mathrm{fit}}}$.
For candidate $c$ and region
$r\in\{1,\ldots,6\}$, a geometric descriptor
$\mathbf g_{rc}$ summarizes observation
support, scaffold and anchor agreement,
fitting and reprojection errors, and ray
geometry. Where enough alternative views
are available, it also evaluates agreement
with a skeleton triangulated without the
candidate's source view, preventing a
candidate from being assessed solely using
evidence it helped establish.
A visual descriptor $\mathbf f_{rc}$
compares features sampled at projected
candidate and anchor joints, including
their differences and image-space
separation, with masked pooling over
available views.
A shared embedding combines both descriptors
with region and candidate indicators.
A scoring network then compares each
candidate with pooled alternatives to predict
a regional score $\ell_{rc}$.
The allocation weights and fused mesh are
\begin{equation}
\begin{aligned}
q_{rc}
&=\underset{c\in\mathcal C}{\operatorname{softmax}}
\left(\log q^0_{rc}+\ell_{rc}\right),\\
\mathbf V_m^{\mathrm{fus}}
&=\mathbf A_m+
\sum_{r=1}^{6}b_{mr}
\sum_{c\in\mathcal C}
q_{rc}(\mathbf C_{cm}-\mathbf A_m),
\end{aligned}
\label{eq:fusion}
\end{equation}
where $q^0_{r,\mathbf A}=0.9$ and
$q^0_{r,v}=0.1/|\mathcal V_{\mathrm{fit}}|$
form an anchor-favoring prior,
$\mathbf C_{cm}$ is vertex $m$ of candidate
$c$, and $b_{mr}$ is a fixed anatomical
basis weight. Invalid candidates are masked,
and the score head is zero-initialized,
so fusion initially remains close to the
anchor and shifts weight toward alternative
candidates where regional evidence supports them.
The three weighting mechanisms serve distinct
purposes: $w_{vj}$ controls the contribution
of joint observations to the shared geometry,
$\pi_v$ combines observation support and
fitting agreement to weight whole-mesh
candidates, and $q_{rc}$ allocates surface
candidates separately across anatomical regions.
Thus, a view can contribute to metric alignment
without being preferred for every part of
the reconstructed surface. The anchor-favoring
prior also provides a common starting point
from which regional scores can adjust the
candidate mixture. Regional fusion is trained
with $\mathcal L_{\mathrm{fus}}=E_V+0.5E_J$,
where $E_V$ and $E_J$ denote the mean
vertex and mesh-derived joint errors
in millimeters without target-based
alignment. The same joint operator
$\mathcal J$ is used for predictions
and targets.

\vspace{0.2cm}
\noindent\textbf{Pixel-aligned refinement.}
Regional fusion integrates geometry already
present in the aligned candidate bank, but
errors shared by those candidates can remain
in the fused surface.
The original images provide additional local
evidence that can be sampled at projected
surface locations, as demonstrated by
pixel-aligned reconstruction and refinement~\cite{saito2019pifu,zhang2021pymaf}.
We use this evidence to predict bounded
residual displacements of the fused surface.

Specifically, we project 195 sampled
vertices of $\mathbf V^{\mathrm{fus}}$
into the available views and sample
features from $\Phi_v$.
Each view token encodes ray geometry,
ray--mesh intersection validity, and
the depth difference between the fused
vertex and the view's aligned candidate
$\widetilde{\mathbf V}_v$ along the pixel
ray, normalized by 1\,cm.
For each sampled vertex, the nearest joint
in $\mathbf X_{\mathrm{geo}}$ identifies
the corresponding 2D correction magnitude
and view-specific support $w_{vj}$.
View attention adds the logarithm of this
support as a score bias and masks invalid
projections. The depth discrepancy describes
how each aligned candidate differs locally
from the fused reconstruction, while the
nearest-joint support carries the reliability
of the geometric observations into the
final surface correction.
Two transformer layers then exchange
information among sampled vertices.
For token $\mathbf h_q$, a zero-initialized
head predicts
$\boldsymbol\delta_q
=0.005\tanh f_{\mathrm{off}}(\mathbf h_q)$
in meters, interpolated to the full mesh:
\begin{equation}
\widehat{\mathbf V}_m
=\mathbf V_m^{\mathrm{fus}}
+\gamma\sum_{q=1}^{195}U_{mq}\boldsymbol\delta_q,
\label{eq:refinement}
\end{equation}
where $U\in\R^{778\times195}$ is a fixed
three-neighbor interpolation matrix
with nonnegative, normalized rows.
The shared model uses $\gamma=1$ in
training and evaluation, bounding each
final displacement coordinate by 5\,mm.
The refinement objective combines
surface and joint supervision with
image reprojection and displacement
regularization:
\begin{equation}
\mathcal L_{\mathrm{ref}}
=E_V+0.5E_J+0.1E_{\mathrm{2D}}+0.01E_{\mathrm{Lap}},
\label{eq:ref_loss}
\end{equation}
where $E_{\mathrm{2D}}$ is the mean joint
reprojection error in crop pixels and
$E_{\mathrm{Lap}}$ penalizes the
mesh-Laplacian norm of the interpolated
displacements.

\section{Experiments}
\label{sec:experiments}

\subsection{Experimental Setup}
\label{sec:datasets}

\vspace{0.1cm}
\noindent\textbf{Datasets.}
We evaluate on DexYCB-MV (eight views, 4,951
right-hand S0 test groups)~\cite{chao2021dexycb},
HanCo (eight views, 20,198 test frames)~\cite{zimmermann2021hanco},
and two five-view HO3D test
populations~\cite{hampali2020honnotate,hampali2021ho3dv3}.
HO3D-Eval-MV contains 3,178 groups from camera
sequences outside the official training split
used by HaMeR~\cite{pavlakos2024hamer},
whereas HO3D-MV contains 2,702 valid HaMeR
evaluation groups with camera-sequence overlap.
Development analyses use 1,412 DexYCB and
3,904 HO3D groups, the latter from official
HO3D training sequences.

\vspace{0.1cm}
\noindent\textbf{Evaluation.}
We use calibrated cameras and annotation-assisted
hand localization. We report mean per-vertex
position error (MPVPE) and mean per-joint
position error (MPJPE) over 21 joints, in millimeters.
DexYCB and HO3D use the absolute reference-camera
frame. HanCo errors are computed after root
translation, without rotation or scale
alignment, as in~\cite{han2026usthand}.

\vspace{0.1cm}
\noindent\textbf{Compared methods.}
We compare with POEM, POEMv2, JUMP-Hand, and
UST-Hand using published results, and additionally
reevaluate the released POEMv2-large weights
on both HO3D test populations.

\vspace{0.1cm}
\noindent\textbf{Implementation.}
We freeze HaMeR and train the downstream modules
sequentially with AdamW~\cite{loshchilov2019adamw}.
The integration model has 0.98M trainable parameters
and shares its weights and hyperparameters across
all three datasets, with $\gamma=1$ in training
and evaluation. Optimization and preprocessing
details are provided in Appendix~\ref{sec:s_method}.

\subsection{Comparison with Existing Methods}
\label{sec:comparison}

\begin{table*}[t]
\centering
\footnotesize
\setlength{\tabcolsep}{2.4pt}
\begin{tabular}{@{}llclrrrrrrrr@{}}
\toprule
& & & &
\multicolumn{2}{c}{DexYCB-MV (8 views)} &
\multicolumn{2}{c}{HanCo$^{\S}$ (8 views)} &
\multicolumn{2}{c}{HO3D-Eval-MV (5 views)} &
\multicolumn{2}{c}{HO3D-MV$^{\ddagger}$ (5 views)} \\
\cmidrule(lr){5-6}
\cmidrule(lr){7-8}
\cmidrule(lr){9-10}
\cmidrule(l){11-12}
Method & Sup. & Frames & Training data &
MPVPE & MPJPE & MPVPE & MPJPE &
MPVPE & MPJPE & MPVPE & MPJPE \\
\midrule
POEM$^{*}$~\cite{yang2023poem}
& Full & 1 & Single
& 6.13 & 6.06 & -- & -- & -- & -- & -- & -- \\
POEMv2-large$^{*}$~\cite{yang2025poemv2}
& Full & 1 & Multi
& 6.50 & 6.33 & -- & -- & -- & -- & 7.63 & 7.18 \\
JUMP-Hand$^{*}$~\cite{kuang2026jumphand}
& Full & 1 & Single
& 5.45 & -- & -- & -- & -- & -- & 13.39 & 13.10 \\
JUMP-Hand$^{*}$~\cite{kuang2026jumphand}
& Full & 1 & Multi
& 5.51 & -- & -- & -- & -- & -- & 8.62 & -- \\
UST-Hand$^{*}$~\cite{han2026usthand}
& Self & 1 & Single
& -- & -- & 5.94 & 5.38 & -- & -- & -- & -- \\
UST-Hand$^{*}$~\cite{han2026usthand}
& Self & 5 & Single
& -- & -- & 5.82 & 5.19 & -- & -- & -- & -- \\
\midrule
POEMv2-large$^{\dagger}$~\cite{yang2025poemv2}
& Full & 1 & Multi
& -- & -- & -- & -- & 10.97 & 10.52 & 7.43 & 6.95 \\
HaMeR + geometry (HG)
& Prior & 1 & Prior only
& 5.997 & 5.892 & 7.166 & 6.643
& 11.119 & 10.875 & 5.252 & 5.034 \\
\textbf{\method}
& Full & 1 & Multi
& \textbf{4.193} & \textbf{4.090}
& \textbf{5.083} & \textbf{4.587}
& \textbf{10.233} & \textbf{9.882}
& \textbf{4.625} & \textbf{4.332} \\
\bottomrule
\end{tabular}
\caption{Benchmark MPVPE/MPJPE (mm; lower is better).
$^{*}$Published results; $^{\dagger}$released weights
reevaluated; $^{\S}$HanCo root translation;
$^{\ddagger}$HO3D-MV training-split overlap.
``Prior'' indicates no learned integration;
Frames counts temporal inputs.
\method uses one shared model.
On HO3D-MV, POEMv2-large is evaluated on
2,706 groups, versus 2,702 valid HaMeR
groups for HG and \method.}
\label{tab:main}
\end{table*}

\vspace{0.1cm}
\noindent\textbf{Comparison with prior methods.}
Table~\ref{tab:main} compares \method with
representative multi-view hand reconstruction
methods under their respective evaluation
protocols. \method obtains the lowest
MPVPE and MPJPE among the compared methods
on all three primary benchmarks.
On DexYCB-MV, its MPVPE is below the
reported 5.45\,mm of JUMP-Hand.
On HO3D-Eval-MV, it achieves 10.233\,mm,
compared with 10.97\,mm for the released
POEMv2-large model on the same 3,178 groups.
On HanCo, \method also improves over the
reported five-frame UST-Hand result.
These comparisons differ in supervision,
pretraining, and preprocessing; UST-Hand
uses self-supervision, whereas \method
trains supervised integration modules
on frozen monocular predictions.

\vspace{0.1cm}
\noindent\textbf{Controlled comparison.}
The matched HaMeR + geometry (HG) baseline
uses the same frozen predictions, crops,
camera calibration, and mesh-to-joint operator
as \method. HG triangulates joints uniformly,
applies the same similarity fitting, and
averages the valid aligned meshes.
Relative to HG, \method reduces MPVPE by
30.1\%, 29.1\%, and 8.0\% on DexYCB-MV,
HanCo, and HO3D-Eval-MV, respectively.

\begin{figure*}[t]
\centering
\includegraphics[width=\textwidth]{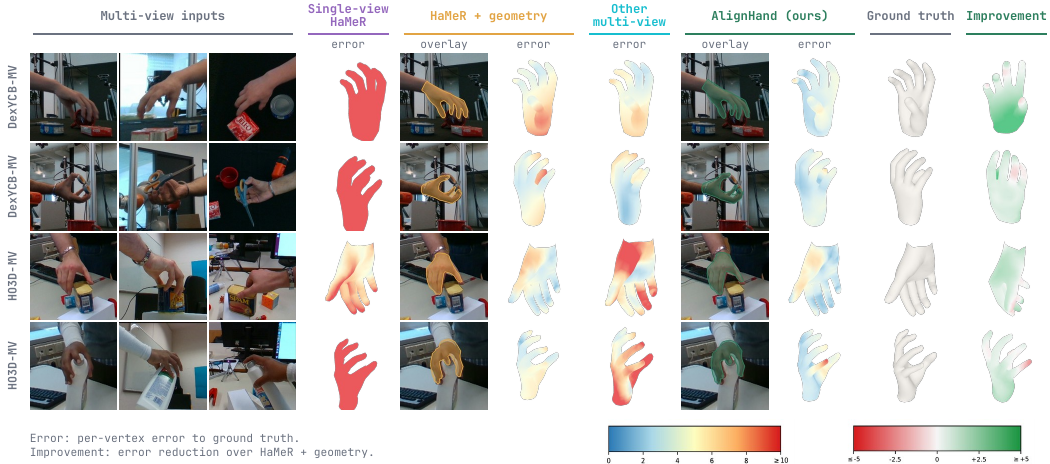}
\caption{\textbf{Qualitative comparison on
DexYCB-MV and HO3D-MV test examples.}
Each row shows three of the synchronized
input views. ``Other multi-view'' is WiLoR +
geometry on DexYCB-MV and POEMv2-large on
HO3D-MV. Errors use a 0--10\,mm scale;
improvement maps show the error reduction
over HG.}
\label{fig:qual_compare}
\end{figure*}

\vspace{0.1cm}
\noindent\textbf{Qualitative comparison.}
Fig.~\ref{fig:qual_compare} illustrates
four examples with different hand poses
and viewing conditions.
Relative to the reference-frame ground truth,
independent monocular predictions exhibit large
absolute reconstruction errors in several examples.
HG reduces these discrepancies through geometric
alignment, while \method further improves
palm and finger regions in the illustrated cases.
The corresponding input overlays show
how these geometric differences appear
when the reconstructed meshes are
reprojected into the images.

\subsection{Ablation and Multi-View Analysis}
\label{sec:ablation}

\begin{table}[t]
\centering
\small
\setlength{\tabcolsep}{5pt}
\begin{tabular}{lrr}
\toprule
Configuration & DexYCB & HO3D \\
\midrule
\multicolumn{3}{l}{\emph{(a) Alignment and enhancement}} \\
Geometry baseline (HG) & 5.567 & 4.161 \\
Aligned anchor & 3.724 & 3.532 \\
+ Regional fusion & 3.545 & \textbf{3.449} \\
+ Pixel-aligned refinement (\method)
& \textbf{3.409} & 3.562 \\
\midrule
\multicolumn{3}{l}{\emph{(b) Observation refinement (anchor only)}} \\
Correction + predicted support
& \textbf{3.724} & \textbf{3.532} \\
No correction; predicted support
& 4.398 & 3.837 \\
Correction; uniform support
& 4.345 & 3.628 \\
No correction; uniform support
& 5.540 & 4.155 \\
\bottomrule
\end{tabular}
\caption{Development MPVPE (mm; lower is better).
(a) Successive reconstruction stages.
(b) Fixed-checkpoint interventions at the anchor;
the final row differs from HG only in anchor weighting.
Bold marks the lowest error per panel and dataset.}
\label{tab:components}
\end{table}

\begin{figure}[t]
\centering
\includegraphics[width=\columnwidth]{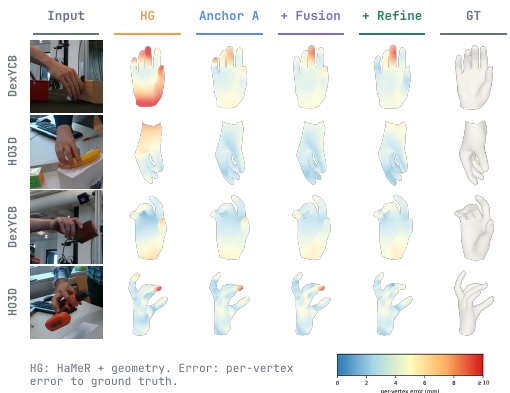}
\caption{\textbf{Stage-wise reconstructions on
DexYCB and HO3D development examples.}
Columns compare HG, aligned anchor,
regional fusion, and pixel-aligned refinement
against ground truth, with vertex errors
shown on a 0--10\,mm scale.}
\label{fig:qual_ablation}
\end{figure}

\vspace{0.1cm}
\noindent\textbf{Alignment and surface enhancement.}
Table~\ref{tab:components}(a) shows that
the observation-guided alignment stage,
including fit-aware anchoring, yields
the largest MPVPE reduction on both
development datasets. On DexYCB,
the anchor accounts for approximately
85\% of the complete pipeline's
improvement relative to HG.
Since HG applies the same similarity fitting,
the gain reflects the improved geometric
reference and fit-aware aggregation.

Fig.~\ref{fig:qual_ablation} illustrates
the corresponding stage-wise changes.
The most prominent reductions in
vertex error occur from HG to the
aligned anchor, while fusion and refinement
produce more localized changes.
On DexYCB, both enhancement stages
further reduce MPVPE. On HO3D, regional fusion reduces MPVPE to
3.449\,mm and Procrustes-aligned vertex
error (PA-V) to 2.464\,mm,
whereas refinement increases them
to 3.562 and 2.512\,mm, respectively.
Regional fusion improves both MPVPE and PA-V
on the two development sets, whereas the
additional effect of pixel-aligned refinement
varies across datasets.

\vspace{0.1cm}
\noindent\textbf{Observation correction and view support.}
Table~\ref{tab:components}(b) evaluates
coordinate correction and predicted support
through fixed-checkpoint interventions
at the aligned anchor.
Disabling coordinate correction or replacing
predicted support with uniform weights
increases anchor MPVPE by 0.674 and
0.621\,mm on DexYCB, and by 0.305 and
0.096\,mm on HO3D, respectively.
Applying both interventions brings anchor
accuracy close to HG despite retaining
fit-aware aggregation. Together with the matched
similarity fitting, this highlights the role of
reliable cross-view observations in establishing
the shared metric reference.
We additionally evaluate final reconstructions
after applying $96\times96$ black patches to
$\lfloor V/2\rfloor$ input views and rerunning
the frozen monocular estimator.
Coordinate correction is retained in both
settings, while uniform weighting replaces
predicted support throughout the pipeline.
Predicted support reduces the corruption-induced
MPVPE increase by 45.4\% on DexYCB and 53.2\%
on HO3D relative to uniform weighting
(Fig.~\ref{fig:s_occlusion}).

\vspace{0.1cm}
\noindent\textbf{Absolute versus aligned errors.}
We compare absolute reconstruction errors
with PA-V, which removes global rotation, translation,
and scale. On DexYCB-MV, \method reduces
absolute MPVPE relative to HG by 1.80\,mm,
while PA-V decreases by only 0.31\,mm.
The corresponding reductions are 2.46 and
0.56\,mm on all HanCo test frames in the
absolute frame, and 0.89 and 0.31\,mm on
HO3D-Eval-MV. On HO3D-MV, absolute MPVPE
decreases by 0.63\,mm despite a 0.20\,mm
increase in PA-V.
The larger absolute improvements are
consistent with a substantial reduction
in global similarity discrepancies.

\begin{figure}[t]
\centering
\includegraphics[width=\columnwidth]{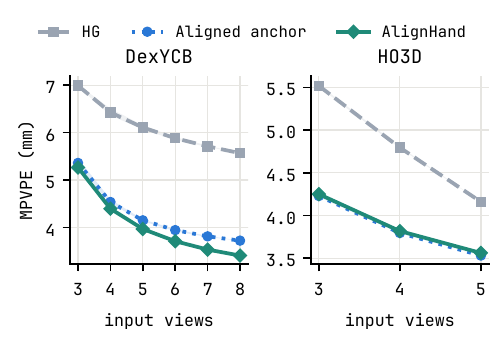}
\caption{Development MPVPE versus input views.
Curves average three nested camera draws;
bands show their range.}
\label{fig:views}
\end{figure}

\vspace{0.1cm}
\noindent\textbf{Additional input views.}
We vary the available camera subset without
retraining the shared model.
As shown in Fig.~\ref{fig:views} and Table~\ref{tab:s_views}, accuracy
improves with additional views on both
development datasets.
On DexYCB, the advantage over HG increases
from 1.724\,mm with three views to 2.158\,mm
with eight, although HG receives the same
additional observations.
Over this range, the improvement from the
aligned anchor to the final mesh also grows
from 0.096 to 0.315\,mm, showing a larger
contribution from surface enhancement in
the evaluated higher-view subsets.
On HO3D, final MPVPE decreases from 4.251
to 3.562\,mm between three and five views;
the anchor benefits similarly, and the two
errors remain close.

\subsection{Shared Model, Prior Compatibility, and Cost}
\label{sec:analysis}

\vspace{0.1cm}
\noindent\textbf{Shared integration across datasets.}
The main results use one integration model
trained on pooled DexYCB, HanCo, and HO3D
data, with shared weights, geometry
normalization, and hyperparameters.
Relative to dataset-adapted configurations,
its MPVPE differs by less than
0.05\,mm on DexYCB-MV and HanCo
and is lower on both HO3D test populations:
10.233 versus 10.750\,mm on HO3D-Eval-MV
and 4.625 versus 4.835\,mm on HO3D-MV
(Table~\ref{tab:s_shared}). 
The same model is
evaluated on eight-view and five-view camera layouts:
relative camera geometry is encoded explicitly,
and the support and candidate weights are normalized
over the available views or candidates.

\vspace{0.1cm}
\noindent\textbf{Alternative monocular prior.}
We also evaluate the framework with frozen
WiLoR~\cite{potamias2025wilor}, retraining
the downstream modules for each prior using
dataset-adapted configurations.
On DexYCB-MV, integration reduces MPVPE
by 1.85\,mm with HaMeR and 1.83\,mm
with WiLoR; on HO3D-Eval-MV, the
corresponding reductions are 0.37 and
0.50\,mm, and HO3D-MV also improves with both priors
(Table~\ref{tab:s_backbone_transfer}).

\vspace{0.1cm}
\noindent\textbf{Runtime.}
At batch size one on a B200 GPU, the eight-view
full pipeline takes 44.01\,ms per group,
including crop warping, normalization, and
host--device transfers, but excluding image
decoding and hand detection.
Standalone HaMeR inference takes 13.10\,ms,
while integration with cached prior outputs
takes 22.25\,ms.
The three settings are measured independently
(Table~\ref{tab:s_shared_runtime}).

\section{Conclusion}
\label{sec:conclusion}

We presented \method, a calibrated multi-view
framework that aligns predictions from a
frozen monocular prior before integrating
their complementary surface evidence.
Our analyses show that reliable cross-view
geometry, through observation-guided alignment
and fit-aware anchoring, provides the largest
and most consistent reconstruction gains.
In the stage-wise analysis, regional fusion
further improves the aligned reconstructions,
while pixel-aligned refinement
has dataset-dependent effects.
With 0.98M trainable parameters and the
monocular estimator unchanged, one shared
integration model improves reconstruction
across three datasets and the evaluated camera
layouts, and the framework remains effective
with another frozen prior after retraining.

{\small
\bibliographystyle{ieeenat_fullname_unsrt}
\bibliography{references}}

\clearpage
\appendix
\setcounter{section}{0}\setcounter{equation}{0}\setcounter{table}{0}\setcounter{figure}{0}
\maketitlesupplementary

\renewcommand{\thesection}{\Alph{section}}
\renewcommand{\theequation}{S\arabic{equation}}
\renewcommand{\thetable}{S\arabic{table}}
\renewcommand{\thefigure}{S\arabic{figure}}

\section{Implementation and Training}
\label{sec:s_method}

\subsection{Module Details}

\paragraph{Cross-view observation refinement.}
Normalized crop coordinates are obtained by
dividing crop-pixel coordinates by 256 and
subtracting $0.5$. Given a reference-to-camera
transform
$T_v=[R_v^{\mathrm{cam}}\mid
\mathbf t_v^{\mathrm{cam}}]$,
a homogeneous raw-image observation
$\overline{\mathbf p}_{vj}$ defines the
camera center, unit ray direction, and
Pl\"ucker moment:
\begin{equation}
\begin{aligned}
\mathbf c_v
&=-(R_v^{\mathrm{cam}})^\top
\mathbf t_v^{\mathrm{cam}},\\
\mathbf d_{vj}
&=
\frac{(R_v^{\mathrm{cam}})^\top
K_v^{-1}\overline{\mathbf p}_{vj}}
{\left\|(R_v^{\mathrm{cam}})^\top
K_v^{-1}\overline{\mathbf p}_{vj}\right\|_2},\\
\mathbf m_{vj}
&=\mathbf c_v\times\mathbf d_{vj}.
\end{aligned}
\label{eq:s_ray}
\end{equation}
These rays are computed from the initial
monocular joint observations, before
coordinate correction.
Pl\"ucker moments use metric camera centers;
the mean-distance normalization below applies
only to the camera-center component of the
relative-pose input.

The observation-refinement network takes
a 17-channel geometric input:
the initial 2D joint coordinates (2),
Pl\"ucker ray direction and moment (6),
and relative camera pose (9).
The pose representation contains the first
two columns of $(R_v^{\mathrm{cam}})^\top$ and the camera center
normalized by the mean distance of the
available cameras from the reference camera.
Appearance features sampled from the
1280-channel HaMeR map and the 17 geometric
channels are projected to 128 channels by
separate linear layers and summed with a
joint-identity embedding.
No camera-identity or physical-view
embedding is used in this module.

Cross-view attention compares observations
of the same joint. Each of four attention
heads has dimension 32, with scores
\begin{equation}
a^h_{jvu}
=
\frac{(\mathbf q^h_{vj})^\top
\mathbf k^h_{uj}}{\sqrt{32}}
-\min\!\left(
\frac{\kappa_{jvu}}{s_j},12
\right),
\end{equation}
where
$\kappa_{jvu}
=|\mathbf d_{vj}^{\top}\mathbf m_{uj}
+\mathbf d_{uj}^{\top}\mathbf m_{vj}|$.
The scale $s_j$ is the median ray
discrepancy over all ordered pairs of
available views, including $u=v$,
floored at $10^{-5}$ and detached
from gradients.
Masked softmax excludes unavailable views.
Residual attention and a
$128\!\rightarrow\!256\!\rightarrow\!128$
GELU feed-forward block update each token;
a single such block is used.
View-availability masks support subsets
of the available calibrated cameras.

The correction head uses
LN$(128)$--Linear$(128)$--GELU--Linear$(2)$,
and the support head uses
LN$(128)$--Linear$(64)$--GELU--Linear$(1)$;
the output layers of both heads are
zero-initialized.
The correction head predicts bounded
corrections in normalized crop coordinates.
Corrected joints are converted to raw-image
pixels for provisional uniform triangulation.
Reprojection errors from that triangulation,
combined with the learned support head,
determine joint-specific view support.

During training, the observation objective
supervises the weighted triangulated
skeleton in 3D and corrected 2D joints
against annotations.
The projection-consistency term
$S_{0.02}^{\mathrm{proj}}$ compares
skeleton reprojections with corrected
joints, not directly with ground-truth
image coordinates.
Both 2D terms are evaluated over
supported annotations.
The entropy penalty is
$\mathcal L_H=\operatorname{mean}_j
\max(0,H_{\min}-H_j)^2$ over joints with
more than one available view, where
$H_j=-\sum_v w_{vj}\log w_{vj}$.
The correction penalty
$\mathcal L_{\Delta u}$ is the mean squared
coordinate correction in normalized crop
coordinates.
In observation refinement, the
reprojection errors $e_{vj}$ and the
provisional uniform triangulation remain
differentiable; only the ray-discrepancy
scale $s_j$ is detached.
In similarity fitting, the Procrustes
rotation is detached, and gradients are
computed with the fitted rotation held
fixed.

\paragraph{Regional fusion.}
The six anatomical joint pools are
palm $(0,1,5,9,13,17)$,
index $(5,6,7,8)$,
middle $(9,10,11,12)$,
little $(17,18,19,20)$,
ring $(13,14,15,16)$,
and thumb $(1,2,3,4)$.
Each region uses six descriptor slots,
with padding for unused digit slots.

The 104 geometric channels consist of
26 summary channels and $6\times13$
joint-slot channels.
The summaries encode support
mean/std/min/max (4);
candidate-to-scaffold mean displacement
and distance mean/std/max (6);
candidate-to-anchor displacement and
distance statistics (6);
retained anchor-distance statistics (3);
placement residual, fusion weight,
and log weight (3);
ray cosine mean/min and camera distance (3);
and candidate validity (1).

Each joint slot contains distances to the
scaffold, retained anchor-joint channels,
and origin-excluded triangulation;
cross-view and origin-view reprojection
errors; anchor-relative bone-length
discrepancy; ray-cosine statistics;
a conditioning ratio; and validity flags.
The origin-excluded ray-system
conditioning ratio is
$\lambda_{\min}(H)/\operatorname{tr}(H)$,
where
\[
H=\sum_{u\ne v}\widetilde w_{uj}
(I-\mathbf r_{uj}\mathbf r_{uj}^{\top}).
\]
Here $\mathbf r_{uj}$ is the unit vector
from camera center $\mathbf c_u$ to
scaffold joint $\mathbf X_{\mathrm{geo},j}$,
and $\widetilde w$ renormalizes support
after excluding the candidate's source view.
Invalid fields are mean-filled before
standardization, with normalization
statistics computed from fitting data.

The regional feature reader maps crop
pixels to
$g_x=2(x-32)/192-1$ and
$g_y=2y/256-1$, respecting HaMeR's
central $192\times256$ backbone footprint.
Bilinear sampling uses
\texttt{align\_corners=False}.
A linear layer projects the 1280-channel
HaMeR features to 16 channels.
Candidate and anchor features, their
difference, raw-pixel separation divided
by 256, and a common-support flag form
$3\times16+2=50$ channels.
For each of the six slots, masked mean
and max pooling over the available views
gives 100 channels; concatenating the
slots produces 600 visual channels.

The regional descriptor combines
104 geometric channels, 600 visual
channels, and 11 indicator channels.
These indicators include a six-dimensional
region one-hot vector, a three-dimensional
candidate-source one-hot vector,
a scalar source-view index, and a
scalar validity flag.
The resulting 715-dimensional descriptor
is processed by
LN$(715)$--Linear$(64)$--SiLU--Linear$(32)$.
The scoring input concatenates this
32-dimensional embedding, the mean and
maximum embeddings of the other valid
candidates for the same region, and the
11 indicators ($3\times32+11=107$), followed
by LN$(107)$--Linear$(64)$--SiLU--Linear$(1)$.
The score output layer is zero-initialized.
Regional scoring and blending use FP64,
while feature projection uses FP32.

The anatomical basis is a fixed
$778\times6$ template asset assigning
regional weights to corresponding vertices.
Regional fusion uses the basis in the
weighted candidate-blending equation
given in the main paper.

\paragraph{Pixel-aligned refinement.}
The fixed mesh reader applies the
MANO 16-joint regressor, appends
fingertip vertices
$(744,320,443,555,672)$,
and reorders the resulting 21 joints
to the POEM convention.
The same reader is applied to predicted
and target meshes, whereas scaffold
placement uses native HaMeR joints.

Template-only farthest-point sampling
selects 195 vertices, starting from
vertex zero.
Predicted displacements at the 195 sampled
vertices are propagated to all 778 vertices
by three-neighbor inverse-distance interpolation.
Interpolation weights are nonnegative
and normalized per output vertex;
selected vertices retain one-hot rows.
The smoothing operator uses
$L=I-D^{-1}A$ for mesh adjacency $A$.

Residual view tokens sum a linear
projection of the sampled 1280-channel
features and a linear projection of
nine geometry channels:
Pl\"ucker direction and moment (6),
prior-mesh depth difference (1),
ray--mesh intersection validity (1),
and nearest-joint correction magnitude
in crop pixels divided by 256 (1).

For each sampled vertex, the prior-mesh
depth is obtained from the corresponding
view's aligned candidate
$\widetilde{\mathbf V}_v$.
The nearest positive ray--triangle
intersection determines its depth along
the projected pixel ray, accepting
either triangle orientation.
The depth difference between the fused
vertex and this intersection is
normalized by 1\,cm. When the ray misses
the candidate, the depth difference and
the validity channel are set to zero.

Each sampled vertex is assigned the
nearest joint in the triangulated
skeleton $\mathbf X_{\mathrm{geo}}$,
computed per group by Euclidean distance.
The joint supplies both its 2D correction
magnitude and the corresponding
view-specific support $w_{vj}$.
A view-scoring network,
LN--Linear$(128)$--GELU--Linear$(1)$,
produces a logit per view, to which the
logarithm of this support is added.
A softmax over usable views weights the
view tokens; invalid projections
contribute no view evidence.

After view aggregation, two pre-normalized
transformer layers exchange information
among the sampled vertices.
They use 128 channels, four attention heads,
and 512-channel feed-forward blocks.
The offset prediction layer,
Linear$(128\!\rightarrow\!3)$ followed by
the bounded $\tanh$, is initialized
to zero. For the shared model,
$\gamma=1$ and the coordinate-wise
offset bound is $0.005$\,m.
The mesh-Laplacian penalty
$E_{\mathrm{Lap}}$ is the mean Euclidean
norm of $L$ applied to the interpolated
full-mesh displacements, after converting
them from meters to millimeters.

\begin{table}[t]
\centering
\footnotesize
\setlength{\tabcolsep}{3pt}
\begin{tabular}{@{}llr@{}}
\toprule
Module & Layers & Params \\
\midrule
\multirow{4}{*}{\shortstack[l]{Observation\\refinement}}
& Appearance $1280\!\to\!128$; geometry $17\!\to\!128$ & \multirow{4}{*}{327,043} \\
& Joint embedding $21\times128$ & \\
& One attention block (4 heads, FF 256) & \\
& Heads: correction (2), support (1) & \\
\midrule
\multirow{3}{*}{\shortstack[l]{Regional\\fusion}}
& Visual projection $1280\!\to\!16$ & \multirow{3}{*}{77,021} \\
& $\varphi$: LN(715)--64--SiLU--32 & \\
& Score: LN(107)--64--SiLU--1 & \\
\midrule
\multirow{4}{*}{\shortstack[l]{Pixel-aligned\\refinement}}
& Appearance $1280\!\to\!128$; geometry $9\!\to\!128$ & \multirow{4}{*}{579,076} \\
& View score LN(128)--128--GELU--1 & \\
& Two transformer layers (4 heads, FF 512) & \\
& Offset $128\!\to\!3$ & \\
\bottomrule
\end{tabular}
\caption{Learned layers and trainable
parameters of the shared model.}
\label{tab:s_arch}
\end{table}

\subsection{Training Configurations}

The learned components are optimized
sequentially, with HaMeR and previously
trained components frozen at each stage.
Metric alignment has no trainable parameters.
All learned modules use AdamW with weight
decay $10^{-4}$.
HaMeR features of size
$1280\times16\times12$ are cached
for downstream training.

Table~\ref{tab:training} lists the
dataset-adapted and shared configurations.
D, H, and O denote DexYCB, HanCo, and HO3D.
The shared configuration uses one pooled
training run per trainable module, a common
set of regional geometry statistics,
and $\gamma=1$ for every dataset.

\begin{table*}[t]
\centering
\small
\setlength{\tabcolsep}{10pt}
\begin{tabular}{llrrrr}
\toprule
Learned component & Training data &
Learning rate & Effective batch &
Update budget & $\gamma$ \\
\midrule
Observation refinement & D / O
& $7.5\times10^{-5}$ & $4$ ($2\times2$)
& $8000$ / $800$ & -- \\
Observation refinement & H final
& $7.5\times10^{-5}$ & $4$ ($2\times2$)
& $8000$ & -- \\
Regional fusion & D / O
& $3\times10^{-4}$ & $64$
& $1280$ / $634$ & -- \\
Regional fusion & H final
& $3\times10^{-4}$ & $64$
& $1280$ & -- \\
Pixel-aligned refinement & D + H + O$^{a}$
& $3\times10^{-4}$ & $64$ ($8\times8$)
& $3840$ & $0.5$ \\
Pixel-aligned refinement & H final
& $3\times10^{-4}$ & $64$ ($8\times8$)
& $1280$ & $1$ \\
\midrule
Observation refinement (shared) & D + H + O
& $7.5\times10^{-5}$ & $4$
& $16800$ & -- \\
Regional fusion (shared) & D + H + O
& $3\times10^{-4}$ & $64$
& $3194$ & -- \\
Pixel-aligned refinement (shared) & D + H + O
& $3\times10^{-4}$ & $64$
& $3840$ & $1$ \\
\bottomrule
\end{tabular}
\caption{Training settings and update budgets.
D/H/O denote DexYCB/HanCo/HO3D.
$^{a}$This pooled residual checkpoint ($\gamma=0.5$)
is used by the dataset-adapted DexYCB and
HO3D configurations; it is distinct from the
shared configuration.
Parenthesized batches give microbatch
size $\times$ accumulation steps.
Observation refinement uses 400 warmup
updates, cosine decay to 0.1 of the
initial learning rate, and gradient
clipping at 1.0.}
\label{tab:training}
\end{table*}

The shared model uses a predefined update
schedule and the checkpoint at its final
scheduled update, without development-set
checkpoint selection.
Training and checkpoint selection use
fitting data only.
The model has 983,140 trainable parameters (Table~\ref{tab:s_arch}):
327,043 for observation refinement,
77,021 for regional fusion, and
579,076 for pixel-aligned refinement.
The refinement checkpoint also stores
913,318 fixed buffer elements for
downsampling, upsampling, mesh faces,
and Laplacian operations.

\subsection{Dataset-Adapted Configurations and Repetitions}

The dataset-adapted models are auxiliary
configurations, distinct from the shared
model reported in the main benchmark table.
For DexYCB and HO3D with frozen HaMeR,
three residual-refinement initializations
are evaluated with observation refinement
and regional fusion fixed.
For WiLoR, three paired fusion/refinement
random seeds are evaluated with fixed
observation refinement.
HanCo uses one final refit.
These repetitions apply to the specified
components, not independent
end-to-end training runs.

HanCo fitting uses sequences 0--1199.
The fitting segment contains 82,282 groups
before component-specific selection.
Regional fusion and pixel-aligned refinement
use a sequence-stratified 8,192-group bundle.
The dataset-adapted D/O residual configurations
use their predefined $\gamma=0.5$ endpoint,
while HanCo and the shared model use
$\gamma=1$. These coefficients are held
fixed during final evaluation.

\section{Evaluation Protocols}
\label{sec:supp_protocol}

\subsection{Inputs, Populations, and Pretraining}

Each synchronized group contains all available
camera observations of one hand at one instant.
DexYCB uses right-hand S0 populations with
1,412 development and 4,951 test groups.
HanCo fitting uses sequences 0--1199,
while the full test population contains
20,198 labeled frames from sequences
1200--1517.

HO3D development uses 3,904 groups from
official training sequences.
HO3D-Eval-MV contains 1,607 AP1 and
1,571 MPM1 groups with official v3 labels,
giving 3,178 test groups.

HO3D-MV contains 1,030 GPMF1 and
1,676 SB1 groups, or 2,706 groups before
validity filtering.
Four SB1 groups have degenerate
HaMeR reference-view crop boxes,
leaving 2,702 HaMeR-scored groups.
The separate WiLoR evaluation reports
2,706 valid groups.
These native evaluation populations are
therefore not identical, and the
backbone-comparison results should be
interpreted as improvements relative to
each prior's own matched geometry baseline.

Localization is annotation-assisted.
Crops use annotated keypoints on DexYCB,
HanCo, and HO3D-MV, and organizer-provided
boxes on HO3D-Eval-MV.
All boxes are expanded by a factor of 2.5
and resized to $256\times256$.
The reconstruction system reads camera
calibration and supplied hand localization,
but no ground-truth 3D meshes or joints
during inference.

HaMeR pretraining includes HO3D training data,
and HanCo extends FreiHAND recordings.
The geometry baseline, aligned anchor,
fusion output, and final reconstruction
share the same frozen prior and its
pretraining exposure.

\paragraph{HO3D overlap with HaMeR training data.}
Every HO3D camera-sequence identifier was
checked against the official v3 split lists.
The HO3D fitting and development groups
come from the official training split,
with development sequences held out from
downstream training.
Eight of the ten HO3D-MV camera sequences
(GPMF10--14 and SB10, SB12, SB14)
also belong to the official training split.
All ten HO3D-Eval-MV camera sequences
belong to the official evaluation split.

HaMeR's per-frame training membership
is unavailable, so this analysis establishes
camera-sequence overlap rather than
frame-level pretraining membership.
No complete five-view HO3D-MV group
avoids the camera-sequence overlap.
We therefore designate HO3D-Eval-MV
as the primary HO3D evaluation.

\begin{table}[t]
\centering
\footnotesize
\setlength{\tabcolsep}{3.5pt}
\begin{tabular}{lrrrrr}
\toprule
Family & Train views & Groups &
HG & Adapted & Shared \\
\midrule
GPMF1 & 5/5 & 1,030 & 4.264 & 4.100 & 3.725 \\
SB1 & 3/5 & 1,672 & 5.860 & 5.288 & 5.180 \\
AP1 (Eval) & 0/5 & 1,607 & -- & -- & -- \\
MPM1 (Eval) & 0/5 & 1,571 & -- & -- & -- \\
\bottomrule
\end{tabular}
\caption{HO3D camera-sequence overlap with
the official training split and
HO3D-MV MPVPE (mm) by family.
Train views denotes the number of camera
sequences in the official training split.
The SB1 count excludes four groups with
degenerate HaMeR reference-view crops.}
\label{tab:s_overlap}
\end{table}

\subsection{HanCo Evaluation Protocol}

The full HanCo test population contains
all 20,198 labeled frames from the
318 sequences 1200--1517, with eight
views and no validity filtering.
The root-translation protocol, which matches
the published UST-Hand results on HanCo,
translates each predicted and target mesh by
its own middle-finger MCP joint,
without rotation or scale alignment,
in the first-camera coordinate frame.

Joints follow the native MANO regressor
with fingertip vertices
745, 317, 445, 556, and 673
in the UST-Hand joint order.
The full-population results under
this protocol differ from the legacy
1,685-group subset used in additional
error decomposition.
The latter subset uses the absolute
reference-camera frame.

\subsection{Published Baseline Settings}

An asterisk in the main benchmark table
indicates values reported in the
original publication.
POEM and JUMP-Hand use supervised,
single-frame multi-view reconstruction.
POEMv2 is jointly trained on five
multi-view datasets plus FreiHAND,
with its published values taken from
corrected Table IV of
arXiv:2408.10581v3 (15 July 2026).
JUMP-Hand is reported separately for
single-dataset and combined-dataset
training.

UST-Hand uses self-supervision from
offline 2D pseudo-labels.
The main table includes its single-frame
and five-frame configurations.
Its full HanCo test list and root-aligned
metric were confirmed with the authors
and reproduced for our evaluations.

\paragraph{POEMv2-large rerun.}
We evaluate the released POEMv2-large
checkpoint using its original preprocessing.
On HO3D-MV, its original packed-data
loader and scorer yield 7.62\,mm MPVPE,
closely matching the published 7.63\,mm.

For evaluation with our scorer, all five
views are retained.
POEMv2 predicts in the frame of the first
packed camera. For each group, the rigid
transform from that frame to our reference
frame is estimated from the two copies of
the same ground-truth mesh vertices, one
in each frame, and is then applied to the
prediction. The transform contains rotation
and translation only and does not use the
prediction; its mean vertex residual is
0.20\,mm.
Under our scorer, POEMv2-large obtains
7.43\,mm MPVPE on all 2,706 HO3D-MV groups,
compared with 7.62\,mm under its own
scorer; HG and \method are scored on
the 2,702 groups with valid HaMeR crops.

For HO3D-Eval-MV, the released loader
is restricted to AP1 and MPM1,
covering the same 3,178 groups used
for our evaluation.
Its reference coordinates already
match our evaluation frame, so no
additional rigid conversion is applied.

\section{Additional Results}
\label{sec:s_results}

\subsection{Complete Reconstruction Metrics}

We first compare absolute and
Procrustes-aligned reconstruction errors
for the shared model reported in the main
paper. Procrustes alignment removes global
rotation, translation, and scale, allowing
the two metrics to distinguish improvements
associated with metric reconstruction
from those remaining after similarity
alignment.

\begin{table*}[t]
\centering
\small
\setlength{\tabcolsep}{9pt}
\begin{tabular}{lrrrr}
\toprule
& \multicolumn{2}{c}{Absolute MPVPE} &
\multicolumn{2}{c}{PA-V} \\
\cmidrule(lr){2-3}
\cmidrule(l){4-5}
Population & HG & Shared \method &
HG & Shared \method \\
\midrule
DexYCB-MV & 5.997 & 4.193 & 3.174 & 2.864 \\
HanCo & 7.400 & 4.944 & 4.257 & 3.698 \\
HO3D-Eval-MV & 11.119 & 10.233 & 6.707 & 6.394 \\
HO3D-MV & 5.252 & 4.625 & 2.335 & 2.533 \\
\bottomrule
\end{tabular}
\caption{Absolute and Procrustes-aligned
vertex errors (mm) of the shared model
on the test populations. HanCo uses all
20,198 test frames in the absolute frame,
rather than the root-translation protocol
of the main benchmark.}
\label{tab:s_shared_absolute}
\end{table*}

The shared model reduces absolute MPVPE
on all four populations.
These reductions are substantially larger
than the corresponding changes in PA-V.
On HO3D-MV, absolute MPVPE decreases
despite a small increase in PA-V.
This pattern is consistent with a
substantial reduction in global
similarity discrepancies.
Table~\ref{tab:s_shared_stages} lists the
development-set stage outputs of the
shared model. On HO3D, regional fusion
reduces both MPVPE and PA-V, whereas
pixel-aligned refinement increases both.

\begin{table}[t]
\centering
\small
\setlength{\tabcolsep}{5pt}
\begin{tabular}{lrrrr}
\toprule
& \multicolumn{2}{c}{DexYCB (Dev)} &
\multicolumn{2}{c}{HO3D (Dev)} \\
\cmidrule(lr){2-3}\cmidrule(l){4-5}
Output & MPVPE & PA-V & MPVPE & PA-V \\
\midrule
HG & 5.567 & 2.564 & 4.161 & 2.521 \\
Aligned anchor & 3.724 & 2.422 & 3.532 & 2.521 \\
+ Regional fusion & 3.545 & 2.268 & 3.449 & 2.464 \\
+ Refinement & 3.409 & 2.242 & 3.562 & 2.512 \\
\bottomrule
\end{tabular}
\caption{Development stage outputs (mm)
of the shared model.}
\label{tab:s_shared_stages}
\end{table}

Table~\ref{tab:s_full_metrics} separately
reports complete reconstruction endpoints
for the dataset-adapted configurations,
including the aligned anchor, regional
fusion, and pixel-aligned refinement.
These auxiliary configurations use their
own fixed training recipes and are
distinct from the shared model above.

\begin{table*}[t]
\centering
\small
\setlength{\tabcolsep}{8pt}
\begin{tabular}{llrrrr}
\toprule
Population & Output &
MPVPE & MPJPE & Wrist-relative V & PA-V \\
\midrule
DexYCB-MV & HG & 5.997 & 5.892 & 7.890 & 3.174 \\
& Aligned anchor & 4.538 & 4.502 & 5.182 & 3.065 \\
& + Regional fusion & 4.307 & 4.223 & 4.957 & 2.887 \\
& \method & 4.150 & 4.052 & 4.783 & 2.825 \\
\midrule
HanCo & HG & 6.327 & 6.009 & 8.216 & 3.554 \\
& Aligned anchor & 5.836 & 5.535 & 7.714 & 3.408 \\
& + Regional fusion & 5.071 & 4.729 & 6.891 & 3.142 \\
& \method & 4.283 & 3.896 & 5.761 & 3.119 \\
\midrule
HO3D-MV & HG & 5.252 & 5.034 & 6.058 & 2.335 \\
& Aligned anchor & 4.845 & 4.619 & 5.232 & 2.240 \\
& + Regional fusion & 4.858 & 4.599 & 5.216 & 2.478 \\
& \method & 4.835 & 4.563 & 5.174 & 2.505 \\
\midrule
HO3D-Eval-MV & HG & 11.119 & 10.875 & 12.977 & 6.707 \\
& Aligned anchor & 10.952 & 10.735 & 12.296 & 6.692 \\
& + Regional fusion & 10.781 & 10.523 & 12.010 & 6.564 \\
& \method & 10.750 & 10.495 & 11.758 & 6.544 \\
\bottomrule
\end{tabular}
\caption{Dataset-adapted test errors (mm).
HanCo uses the legacy 1,685-group subset;
MPVPE/MPJPE are absolute-frame errors.
Wrist-relative V removes wrist translation;
PA-V applies similarity alignment.
DexYCB/HO3D final results average three
refinement initializations; HanCo uses
one final refit.}
\label{tab:s_full_metrics}
\end{table*}

Metric alignment reduces absolute error
on all four dataset-adapted populations.
Regional fusion provides additional
improvements on DexYCB, HanCo, and
HO3D-Eval-MV, while the effect of
pixel-aligned refinement varies.
On HO3D-MV, the final MPVPE remains
close to that of the anchor, although
mesh-derived joint error decreases.

The HanCo main benchmark uses all
20,198 labeled test frames with the
root-translation protocol.
Table~\ref{tab:s_hanco_full} provides
the corresponding root-translated,
absolute, and Procrustes-aligned
errors on this full population.

\begin{table*}[t]
\centering
\small
\setlength{\tabcolsep}{8pt}
\begin{tabular}{lrrrrrr}
\toprule
& \multicolumn{2}{c}{Root translation} &
\multicolumn{2}{c}{Absolute} &
\multicolumn{2}{c}{Procrustes} \\
\cmidrule(lr){2-3}
\cmidrule(lr){4-5}
\cmidrule(l){6-7}
Model & MPVPE & MPJPE &
MPVPE & MPJPE & PA-V & PA-J \\
\midrule
HG & 7.166 & 6.643 & 7.400 & 7.012 & 4.257 & 4.023 \\
\method (HanCo refit)
& 5.051 & 4.556 & 4.877 & 4.466 & 3.641 & 3.227 \\
\method (shared)
& 5.083 & 4.587 & 4.944 & 4.529 & 3.698 & 3.274 \\
\bottomrule
\end{tabular}
\caption{HanCo errors (mm) on the full
20,198-frame test population under the
root-translation, absolute, and Procrustes protocols.}
\label{tab:s_hanco_full}
\end{table*}

The legacy 1,685-group subset and full
HanCo population have different sizes
and evaluation protocols.
On the subset, absolute MPVPE is
6.327, 4.283, and 4.312\,mm for HG,
the dataset-adapted refit, and the
shared model, respectively.
These values are therefore reported
separately from the full-population
root-translated results.

\subsection{Shared and Dataset-Adapted Models}

The main paper reports the fully shared
integration model. Table~\ref{tab:s_shared}
compares this model with dataset-adapted
configurations under their respective
evaluation protocols.

\begin{table}[t]
\centering
\small
\setlength{\tabcolsep}{4.5pt}
\begin{tabular}{lrrr}
\toprule
Population & Adapted & Shared & Difference \\
\midrule
DexYCB-MV & 4.150 & 4.193 & $+0.043$ \\
HanCo (root translation) & 5.051 & 5.083 & $+0.032$ \\
HO3D-MV & 4.835 & 4.625 & $-0.210$ \\
HO3D-Eval-MV & 10.750 & 10.233 & $-0.517$ \\
\bottomrule
\end{tabular}
\caption{Test MPVPE (mm).
Difference is shared minus adapted.
HanCo uses root translation;
other populations use absolute coordinates.}
\label{tab:s_shared}
\end{table}

The shared model uses 16,800
observation-refinement updates, 3,194 regional-fusion
updates, and 3,840 pixel-aligned-refinement
updates. Dataset-adapted configurations
have distinct training recipes, including
different update budgets and residual
strengths.

Additional development experiments compare
joint pretraining followed by target
fine-tuning against further single-dataset
training updates. These yield 3.370 versus
3.315\,mm on DexYCB and 3.429 versus
3.515\,mm on HO3D.

\subsection{Complete View-Count Results}

Camera views are removed before
observation refinement, with the reference
camera always retained.
The same learned modules process each
available subset. Results average three
nested camera draws; full-view values use
the complete available camera set.

Table~\ref{tab:s_views} reports the
shared-model development results with
three to eight views on DexYCB and
three to five views on HO3D.
The geometric reference and candidate
bank are reconstructed from the available
views for each subset.

\begin{table}[t]
\centering
\small
\setlength{\tabcolsep}{7pt}
\begin{tabular}{lrrrr}
\toprule
Dataset & Views & HG & Anchor & \method \\
\midrule
DexYCB & 3 & 6.989 & 5.361 & 5.265 \\
& 4 & 6.424 & 4.539 & 4.397 \\
& 5 & 6.107 & 4.152 & 3.969 \\
& 6 & 5.882 & 3.946 & 3.711 \\
& 7 & 5.710 & 3.814 & 3.535 \\
& 8 & 5.567 & 3.724 & 3.409 \\
\midrule
HO3D & 3 & 5.517 & 4.229 & 4.251 \\
& 4 & 4.795 & 3.796 & 3.819 \\
& 5 & 4.161 & 3.532 & 3.562 \\
\bottomrule
\end{tabular}
\caption{Development MPVPE (mm)
of the shared model versus view count.}
\label{tab:s_views}
\end{table}

\subsection{Ablation Interventions and Synthetic Occlusion}

The observation-refinement ablations
intervene on the fixed network outputs
without retraining the network.
Removing coordinate correction restores
the initial monocular 2D joints while
retaining predicted view support.
Replacing predicted support with
uniform support retains corrected joints.
The final intervention disables both
coordinate correction and predicted support.

All observation-refinement interventions
are evaluated at the aligned anchor,
before regional fusion and
pixel-aligned refinement.
They retain fit-aware anchor aggregation,
whereas HG averages the geometrically
aligned meshes uniformly.
Thus, the final intervention and HG
differ only in anchor aggregation.

For synthetic occlusion,
$\lfloor V/2\rfloor$ views per group
are corrupted with $96\times96$
black patches: four views on DexYCB
and two on HO3D.
View identities and patch positions
are sampled using fixed seed 20261028
and shared by all compared variants.
Both clean and corrupted inputs rerun
the frozen monocular estimator;
downstream checkpoints remain unchanged.

\begin{figure}[t]
\centering
\includegraphics[width=\columnwidth]{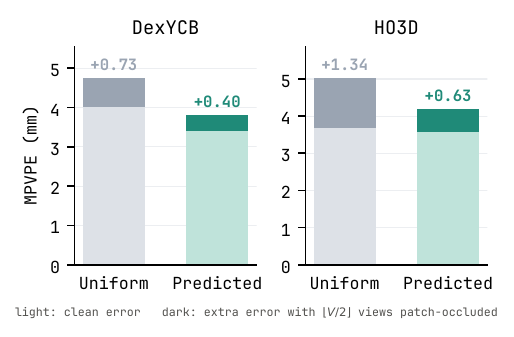}
\caption{Synthetic occlusion on the DexYCB and HO3D
development sets. $V$ denotes the number of available
input views (eight on DexYCB and five on HO3D);
$\lfloor V/2\rfloor$ views are occluded. Final-reconstruction
MPVPE is compared under predicted and uniform support,
with coordinate correction retained in both. Dark
segments show the corruption-induced error increase.}
\label{fig:s_occlusion}
\end{figure}

Unlike the anchor-only interventions,
synthetic occlusion is evaluated on
the final reconstruction.
Coordinate correction remains enabled
in both settings, while uniform weighting
replaces the complete predicted-support
rule throughout the pipeline, including
its geometric reprojection term.
Fig.~\ref{fig:s_occlusion} shows the
resulting errors.

\subsection{Prior Choice and Residual Strength}

Table~\ref{tab:s_backbone_transfer} reports
test-set backbone compatibility using
dataset-adapted configurations.
For each frozen prior, downstream integration
modules are retrained and evaluated against
that prior's own geometry baseline.
Integration improves MPVPE with both HaMeR
and WiLoR across the evaluated test populations.

\begin{table}[t]
\centering
\small
\setlength{\tabcolsep}{5pt}
\begin{tabular}{lrr}
\toprule
Test set & HaMeR & WiLoR \\
\midrule
DexYCB-MV & $5.997\to4.150$ & $6.692\to4.864$ \\
HO3D-Eval-MV & $11.119\to10.750$ & $10.807\to10.308$ \\
HO3D-MV & $5.252\to4.835$ & $5.149\to4.786$ \\
\bottomrule
\end{tabular}
\caption{Test MPVPE (mm): geometry baseline
$\to$ integrated reconstruction, using
dataset-adapted configurations with downstream
modules retrained for each frozen prior.
Integrated values average three runs:
refinement initializations for HaMeR and
paired fusion/refinement initializations
for WiLoR.}
\label{tab:s_backbone_transfer}
\end{table}

Table~\ref{tab:s_prior_gamma} additionally
reports development experiments on
dataset-adapted configurations.
The residual-strength sweep evaluates
fixed trained endpoints.

\begin{table}[t]
\centering
\small
\setlength{\tabcolsep}{10pt}
\begin{tabular}{lrr}
\toprule
Setting & DexYCB & HO3D \\
\midrule
HaMeR prior, complete & 3.364 & 3.479 \\
WiLoR prior, complete & 3.941 & 3.697 \\
\midrule
$\gamma=0$ & 3.472 & 3.523 \\
$\gamma=0.25$ & 3.400 & 3.479 \\
$\gamma=0.5$ & 3.364 & 3.479 \\
$\gamma=0.75$ & 3.367 & 3.521 \\
$\gamma=1$ & 3.408 & 3.603 \\
\bottomrule
\end{tabular}
\caption{Development MPVPE (mm)
of dataset-adapted configurations.
The $\gamma$ sweep evaluates fixed
trained endpoints.}
\label{tab:s_prior_gamma}
\end{table}

Under the evaluated recipes, HaMeR
produces lower development MPVPE
than WiLoR. The residual sweep favors
$\gamma=0.5$ on DexYCB, while
$\gamma=0.25$ and $\gamma=0.5$
both give 3.479\,mm on HO3D.
This sweep characterizes sensitivity
to residual strength in the
dataset-adapted configurations.

The dataset-adapted D/O models retain
their predefined $\gamma=0.5$ blend.
HanCo uses its predefined unshrunk
residual output at $\gamma=1$.
The fully shared deployment model
uses $\gamma=1$ consistently
across all datasets.

\subsection{Runtime Measurement}

The shared-stack benchmark is performed
on one B200 GPU with batch size one
per synchronized group, 20 warmup
iterations, and 200 synchronized
measurement iterations.
All valid camera views are batched
for HaMeR inference.
Neural-network computation primarily
uses FP32; geometric operations and
regional scoring use FP64; cached
HaMeR features are stored in FP16.

The uncached measurement begins with
pre-decoded frames and supplied hand
boxes, and ends with the output mesh
transferred to the host. It includes
crop warping, normalization, device
transfers, frozen HaMeR inference,
and downstream integration.
Image decoding and hand detection
are excluded.

\begin{table}[t]
\centering
\footnotesize
\setlength{\tabcolsep}{4pt}
\begin{tabular}{lrr}
\toprule
Operation (ms/group) & DexYCB (8v) & HO3D (5v) \\
\midrule
\multicolumn{3}{l}{\emph{Downstream, cached prior}} \\
Observation refinement & 3.63 & 3.41 \\
Metric alignment & 2.55 & 2.42 \\
Candidate assembly & 10.08 & 8.06 \\
Regional fusion & 2.58 & 2.13 \\
Pixel-aligned refinement & 3.41 & 3.11 \\
Total & 22.25 & 19.12 \\
\midrule
\multicolumn{3}{l}{\emph{End to end, uncached}} \\
Crop warping & 7.46 & 4.73 \\
Normalization and transfer & 0.94 & 0.67 \\
HaMeR & 13.40 & 11.63 \\
Downstream & 22.16 & 19.83 \\
Output transfer & 0.05 & 0.05 \\
Total & 44.01 & 36.90 \\
\bottomrule
\end{tabular}
\caption{Component timings of the shared
stack (ms/group). Cached and uncached
sections are measured separately.
Totals are computed from unrounded
measurements.}
\label{tab:s_shared_runtime}
\end{table}

\paragraph{Execution settings.}
Standalone frozen HaMeR inference takes
13.10/11.30\,ms for eight/five views,
while integration on cached prior outputs
takes 22.25/19.12\,ms.
The complete uncached pipeline takes
44.01/36.90\,ms, including crop
preprocessing, model inference, and
device transfers.
Table~\ref{tab:s_shared_runtime} reports
the independently measured component
timings; the uncached measurement includes
its own HaMeR and downstream timings.

In the same measurement session,
the dataset-adapted chain takes
22.04/19.25\,ms downstream and
44.58/35.67\,ms end to end on
DexYCB/HO3D.

\begin{table}[t]
\centering
\small
\setlength{\tabcolsep}{4pt}
\begin{tabular}{lrrrr}
\toprule
& \multicolumn{3}{c}{CPU cache-to-mesh (ms)} & GPU (ms) \\
Dataset & Mean & Median & P95 & Batch 1 \\
\midrule
DexYCB & 104.3 & 107.1 & 120.8 & 18.53 \\
HanCo & 95.7 & 93.3 & 105.4 & 18.51 \\
HO3D & 68.7 & 70.6 & 82.9 & 16.98 \\
\bottomrule
\end{tabular}
\caption{Earlier downstream timing of
dataset-adapted configurations.
CPU uses one PyTorch thread;
GPU uses a B200 after warmup.
Prior image inference is excluded.}
\label{tab:s_cpu_timing}
\end{table}

In the earlier benchmark,
observation refinement, metric alignment,
candidate assembly, regional scoring,
and pixel-aligned refinement take
2.93/2.20/8.68/1.97/2.75\,ms on DexYCB,
2.92/2.23/8.67/1.95/2.73\,ms on HanCo,
and 3.00/2.24/7.27/1.80/2.68\,ms on HO3D.
Candidate assembly is the largest
measured downstream component in both
sessions.

\subsection{Reconstruction with Estimated Cameras}

As an exploratory experiment, we evaluate whether the hand-reconstruction
architecture can operate with camera parameters
estimated by external scene-geometry models.
VGGT-1B, $\pi^3$, and Depth Anything 3
provide estimated intrinsics and extrinsics;
their depth predictions are not directly
used by the hand reconstruction modules.

For $\pi^3$, pinhole intrinsics are fitted
from the estimator's own local point maps.
For all three estimators, camera translations
are scaled using the median ratio between frozen-prior
bone lengths and lengths triangulated
under the estimated cameras.
This scale bridge uses neither annotated
bone lengths nor calibrated camera baselines.

All 1,412 DexYCB and 3,904 HO3D
development groups are evaluated
in the reported estimated-camera settings.
These configurations use per-dataset
pixel-aligned refinement with $\gamma=1$
and fixed fitting statistics.
The calibrated reference columns use
the corresponding dataset-adapted
development settings.

\begin{table}[t]
\centering
\small
\setlength{\tabcolsep}{4pt}
\begin{tabular}{llrrr}
\toprule
Dataset & \shortstack[c]{Estimated\\cameras} &
\method & \shortstack[c]{HG\\(calib.)} &
\shortstack[c]{Adapted\\(calib.)} \\
\midrule
DexYCB & DA3 & 2.320 & 2.564 & 2.209 \\
& $\pi^3$ & 2.366 & 2.564 & 2.209 \\
& VGGT-1B & 2.300 & 2.564 & 2.209 \\
HO3D & VGGT-1B & 2.595 & 2.521 & 2.479 \\
\bottomrule
\end{tabular}
\caption{Estimated-camera development PA-V
(mm; lower is better). Calibrated references
use ground-truth cameras; all errors are
similarity-aligned.}
\label{tab:s_cameras}
\end{table}

These experiments use PA-V, which
removes global similarity differences,
and therefore do not independently
establish metric-scale accuracy from
the estimated cameras.
On DexYCB, the VGGT configuration
obtains 2.300\,mm PA-V versus
2.564\,mm for calibrated HG.
On HO3D, the corresponding values
are 2.595 and 2.521\,mm.
The results demonstrate operation
with externally estimated cameras,
while the primary manuscript setting
remains calibrated multi-view reconstruction.
Camera-estimation runtime is excluded
from the calibrated inference benchmark.

\end{document}